\documentclass[a4paper, 10pt, conference]{ieeeconf}
\IEEEoverridecommandlockouts
\usepackage[cmex10]{amsmath}
\usepackage{amssymb}
\usepackage{amsfonts}
\usepackage{algorithm}
\usepackage{algpseudocode}
\usepackage{amsthm}
\usepackage{graphicx}
\usepackage{epsfig}
\usepackage{cite}
\usepackage{tensor}
\usepackage{color}
\usepackage{booktabs} 
\usepackage{hyperref}
\usepackage{geometry}

\hypersetup{colorlinks=true,linkcolor=blue,filecolor=blue,urlcolor=red}

\graphicspath{{figures/}}
\DeclareGraphicsExtensions{.eps}

\theoremstyle{definition}

\title{\LARGE \bf A Point Cloud-Based Method for Automatic Groove Detection and Trajectory Generation of Robotic Arc Welding Tasks}

\author{Rui Peng, David Navarro-Alarcon, Victor Wu and Wen Yang
\thanks{This work is supported in part by the Chinese National Engineering Research Centre for Steel Construction (Hong Kong Branch) at The Hong Kong Polytechnic University under grant BBV8, in part by the Research Grants Council of Hong Kong under grant 142039/17E and in part by The Hong Kong Polytechnic University under grant G-YBYT. 
}
\thanks{RP, DNA, and VW are with The Hong Kong Polytechnic University, Kowloon, Hong Kong. 
Corresponding author: {\texttt{\small dna@ieee.org}}}%
\thanks{WY is with the Wuhan University, Hubei, People's Republic of China.}%
}

\begin{document}
\maketitle
\bstctlcite{IEEEexample:BSTcontrol}
\thispagestyle{empty}
\pagestyle{empty}

\begin{abstract}
In this paper, in order to pursue high-efficiency robotic arc welding tasks, we propose a method based on point cloud acquired by an RGB-D sensor.
The method consists of two parts: welding groove detection and 3D welding trajectory generation. 
The actual welding scene could be displayed in 3D point cloud format. 
Focusing on the geometric feature of the welding groove, the detection algorithm is capable of adapting well to different welding workpieces with a V-type welding groove. 
Meanwhile, a 3D welding trajectory involving 6-DOF poses of the welding groove for robotic manipulator motion is generated. 
With an acceptable error in trajectory generation, the robotic manipulator could drive the welding torch to follow the trajectory and execute welding tasks. In this paper, details of the integrated robotic system are also presented. 
Experimental results prove application value of the presented welding robotic system.
\end{abstract}

\section{INTRODUCTION}\label{sec:intro}
Nowadays, industrial robotic manipulators are used extensively in many factories around the world, with the teach-playback method dominating the robotic welding field.
More specifically, in order to manipulate a robotic arm for welding tasks, a human operator needs to set every path point with precise 3D positioning and 3D orientation in advance. 
However, although CAD-based approaches \cite{norberto2004cad} can generate rather accurate welding trajectories, they still require a great deal of human intervention, as with teach-playback method. 
These conventional robotic welding applications are already unable to cope with the growing welding demand of the construction industry.

To improve robotic welding efficiency, sensors are considered to assist manipulators in automatically locating the welding groove. 
Prevailing sensors involve vision sensors \cite{li2009measurement, liu2014iterative, diao2017passive}, 
RGB-D sensors \cite{li2016new, rodriguez2017feasibility, ahmed2016object}, 
infrared sensors
and laser sensors.  
But laser sensors are much more expensive than vision sensors. 
To control cost in research, vision sensors are more appealing.
During actual robotic welding tasks,   
it is essential to plan a high-precise trajectory \cite{ma2010robot} of the welding groove between two workpieces to achieve acceptable welding quality in complex and unpredictable situations.
Recently, 2D imaging-based methods have become useful in assisting robotic welding in various industrial environments \cite{rao2018real}. 
But with respect to common welding tasks, 2D vision sensors could capture only one frame image of the workpieces. 
An image acquisition system using a CCD video camera, is established by L. Nele et al. \cite{nele2013image}  for real-time weld seam tracking.

However, 2D imaging processing algorithms, which rely on color information, cannot deal with dramatic environmental brightness variation, particularly in the welding fields.
Researchers prefer 3D vision sensors which provide abundant information of the welding environment.
Furthermore, with advancement of the Point Cloud Library (PCL) \cite{rusu20113d}-specifically designed for 3D point cloud processing, 
it is possible to extract and locate the welding seam region in the surface point cloud of welding workpieces. 
The stereo camera as a 3D sensor, is used to realize welding seam reconstruction and tracking \cite{xu2017welding}. The tracking and planning accuracy is good for welding requirements.
An unorganized point cloud-based edge and corner recognition approach \cite{ahmed2018edge} is proved for applicable robotic welding. Also the algorithm has accuracy beyond some of relevant algorithms on 3D point cloud processing. 
Some methods which combine depth data and RGB images have been proposed \cite{song2014sliding}. These methods implement low-cost 3D sensors (e.g. Intel RealSense) \cite{yang2018pixor}.
Through integrating an RGB-D sensor into the robotic system \cite{maiolino2017flexible}, 
the controller is able to cope with the dynamic welding environment. 
Li et al. \cite{jing2016rgb} proposed a welding groove detection approach by an RGB-D sensor.
The approach employs RGB images to recognize the weld groove and a point cloud  to acquire the pose of the targeted weld groove.
 
One of the main problems in the aforementioned research works based on 3D sensors is that their experimental results lack sufficient testing of different types of welding workpieces; 
moreover, the targeted welding groove is simple. 
Inspired by \cite{patil2019extraction}, the proposed method in this paper focuses on welding groove detection and 3D motion trajectory generation. 
Experimental results involve three aspects: runtime efficiency, groove detection accuracy, and trajectory execution. 
The resulting performance of the system on four types of workpieces with V-type grooves proves feasibility in actual welding applications.

\vspace{2pt}
The main contributions of this paper are as follows:
\begin{enumerate}
    \item Develop an integrated intelligent robotic system to automatically execute welding tasks without much human intervention.
    \item Propose a point cloud based welding groove detection algorithm for unpredictable workpieces.
    \item Implement the automatic 3D welding trajectory generation method on a 6-DOF robotic arm.
\end{enumerate}

\section{ROBOT SYSTEM IMPLEMENTATION}\label{sec:Robot System Setup}

\begin{figure}[htbp]
\centering
\includegraphics[width=\columnwidth, height=4cm]{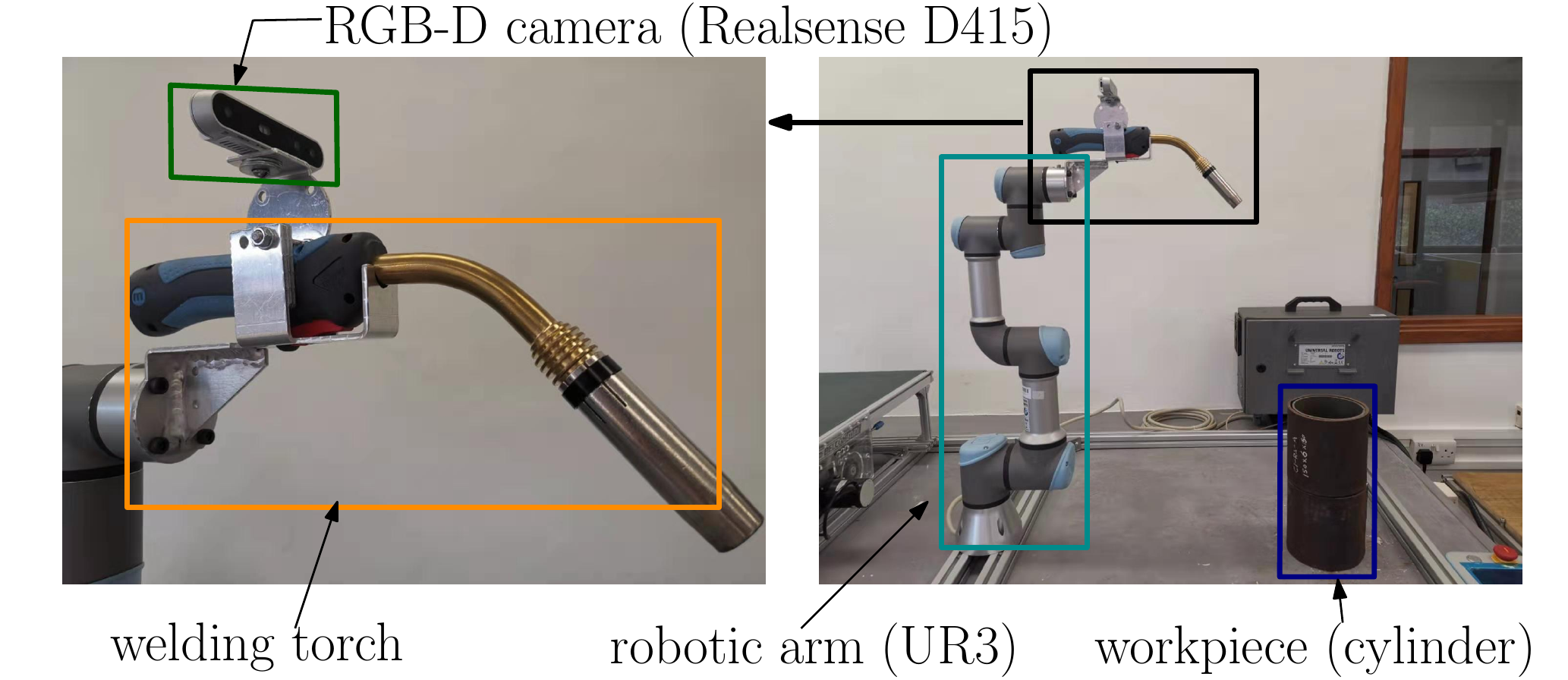}
\caption{(Left) End-effector of the proposed welding robotic system comprises an RGB-D camera and welding torch. 
          The camera is situated above the welding torch. 
         (Right) Experimental welding platform.}
\label{fig:setup}
\end{figure}

The integrated robotic system contains a robotic manipulator (Universal Robot 3), 
an RGB-D camera (Intel Realsense D415), 
and a welding torch. 
The end-effector and whole experimental platform are shown in Fig. \ref{fig:setup}. 
The welding torch is tightly installed on the end-wrist of the robotic manipulator by a supportive metal structure aimed at stable welding execution. 
In addition, the RGB-D camera is physically connected to the welding torch instead of at a fixed position within the welding platform. 
The RGB-D camera moves with the welding torch. 
This design enables the system to flexibly cope with unpredictable welding situations. 
It is also convenient for hand-eye calibration to obtain the accurate transformation matrix from the camera coordinate to the UR3 base coordinate.

\begin{figure}[htbp]
\centering
\includegraphics[width=\columnwidth, height=4cm]{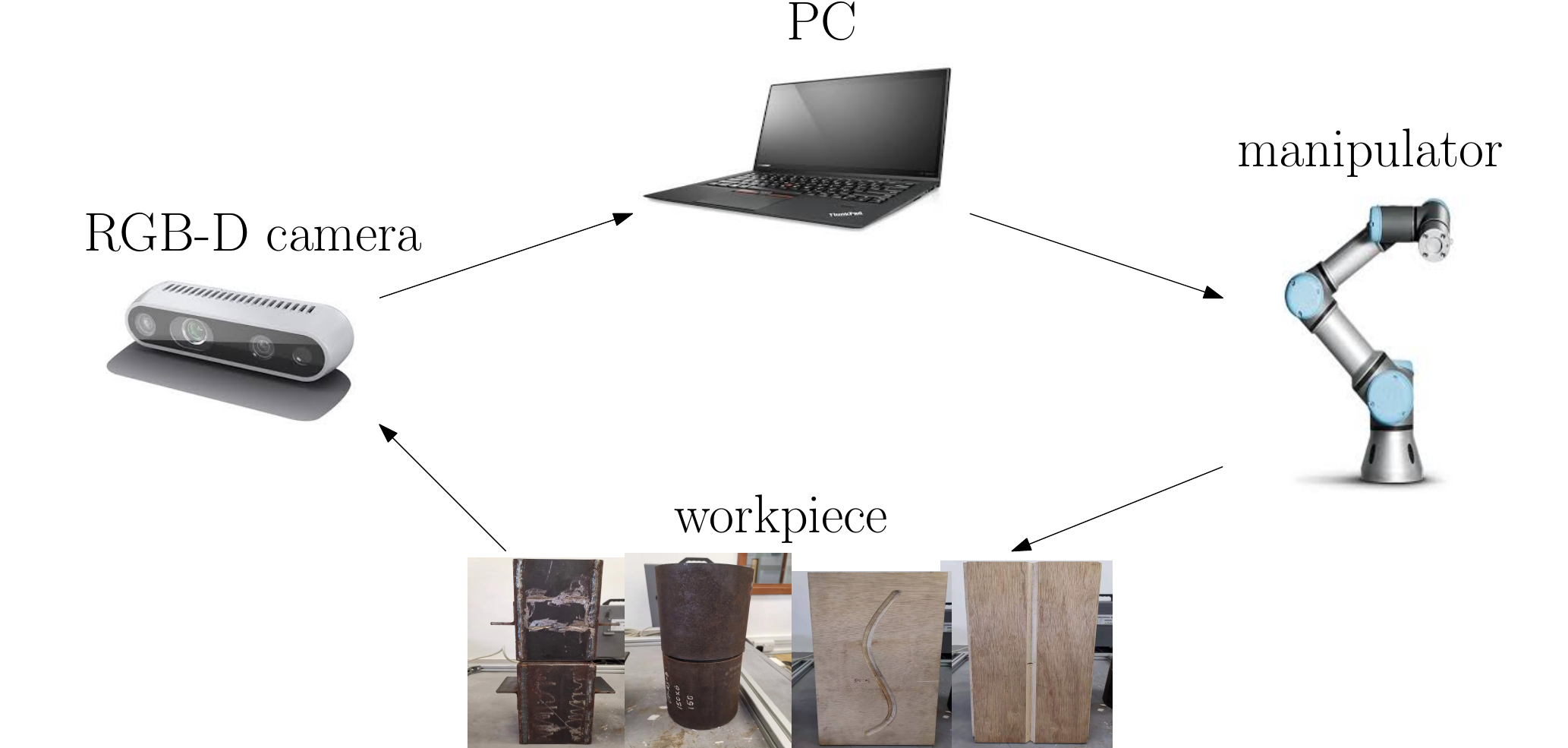}
\caption{Workflow of the proposed robotic system: 
        (1) the RGB-D camera captures the surface point cloud of the workpiece; 
        (2) the welding groove detection algorithm locates the welding groove region in the input point cloud; 
        (3) the trajectory generation method processes the groove point set and outputs a 3D welding trajectory; 
        (4) the manipulator automatically executes the welding motion while tracking the generated trajectory.}
\label{fig:workflow}
\end{figure}

The system running procedure is shown in Fig. \ref{fig:workflow} 
in which the PC with ubuntu 16.04 is a Lenovo-Thinkpad whose CPU uses Intel i5. 
ROS \cite{quigley2009ros}, An open-source robot operating system is used to build a software framework that makes it convenient to develop algorithm modules.
The welding groove detection algorithm completely relies on PCL to process the point cloud and extract the geometric feature. 
The Moveit \cite{chitta2012moveit} package, as one part of the software system, is used to control the UR3 robotic arm with motion planning and collision avoidance.

\section{WELDING GROOVE DETECTION}\label{sec:DETECTION}
\begin{figure}[htbp]
    \centering
    \includegraphics[width=\columnwidth, height=1.5cm]{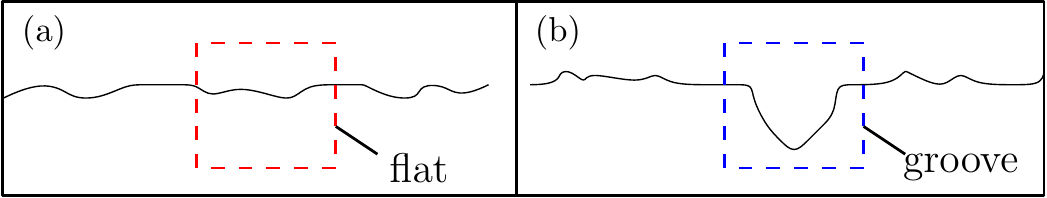}
    \caption{Geometric feature diagram of surface. (a) Flat region. (b) Groove region.}
    \label{fig:geometric_feature}
\end{figure}

The welding groove detection algorithm extracts the groove region by computing the geometric feature of the input point cloud, 
which represents the surface profile of the welding workpiece. 
The geometric feature is defined as the surface variation. 
In other words, the extent of surface slope change. 
Compared with flat and smooth regions, the groove region has higher surface variation (see Fig. \ref{fig:geometric_feature}).

In order to mathematically describe the surface variation, 
an efficient 3D feature histogram in the algorithm was designed. 
Furthermore, a surface variation descriptor is computed for each point of the input point cloud. By setting a threshold value for the descriptor, 
the groove region can be separated from other regions.

\subsection{Input Point Cloud Preprocessing}
\begin{figure}[htbp]
    \centering
    \includegraphics[width=\columnwidth, height=6cm]{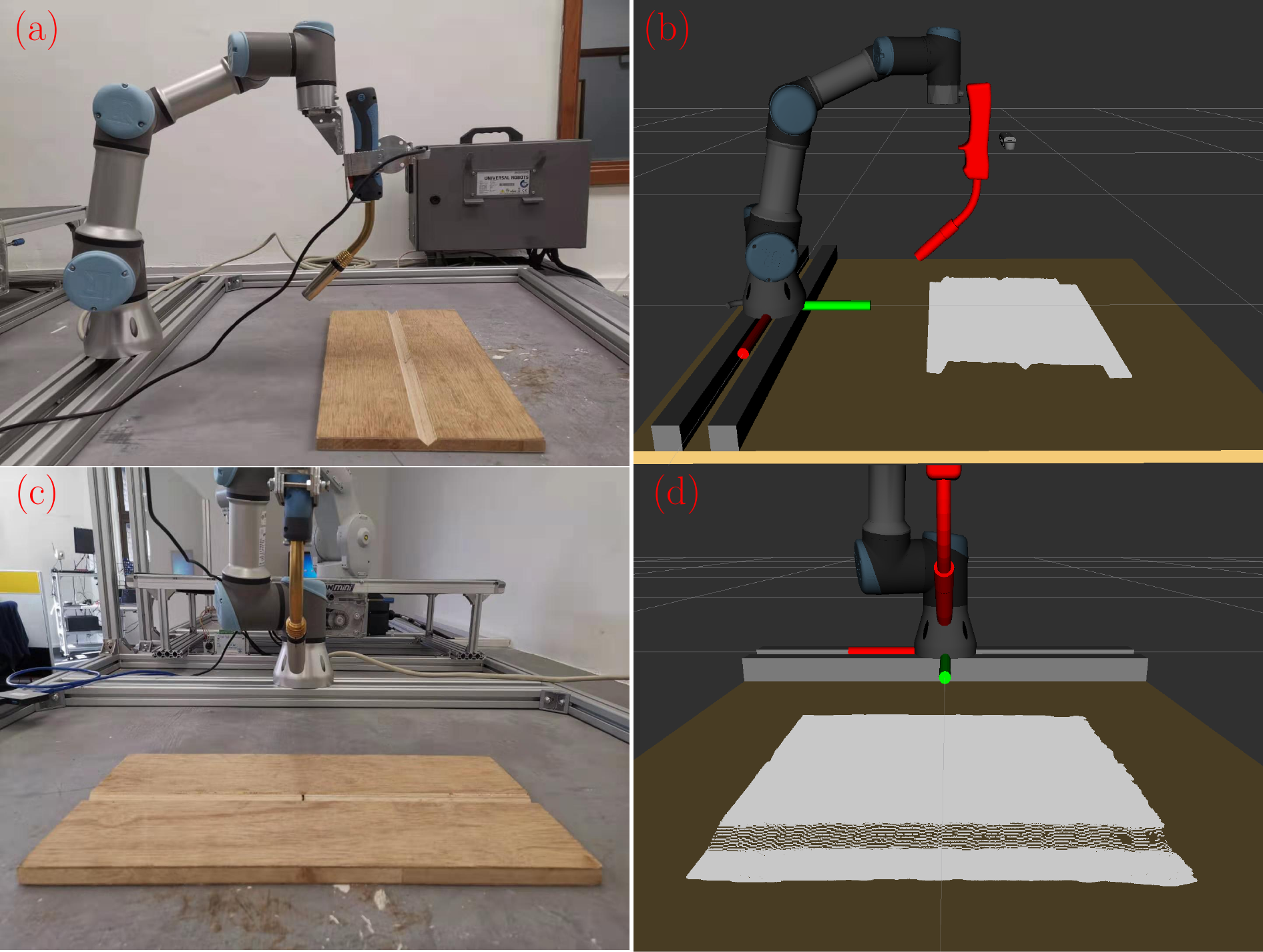}
    \caption{The RGB-D camera captures the surface point cloud of the workpiece with a straight-line welding groove. 
            (a) A robotic manipulator moves so that the camera can capture the complete point cloud. 
            (b) The surface point cloud is shown in the simulation (Rviz). 
            (c) The actual workpiece is viewed from the side. 
            (d) The raw point cloud is organized.}
    \label{fig:captures}
\end{figure}
    
First, 
the RGB-D camera attached to the robotic arm moves to an appropriate position and captures one frame of a raw point cloud which covers the whole surface of the workpiece, 
as shown as Fig. \ref{fig:captures}.
Then the raw point cloud is taken as an input of the welding groove detection algorithm.

\begin{figure}[htbp]
\centering
\includegraphics[width=\columnwidth, height=3cm]{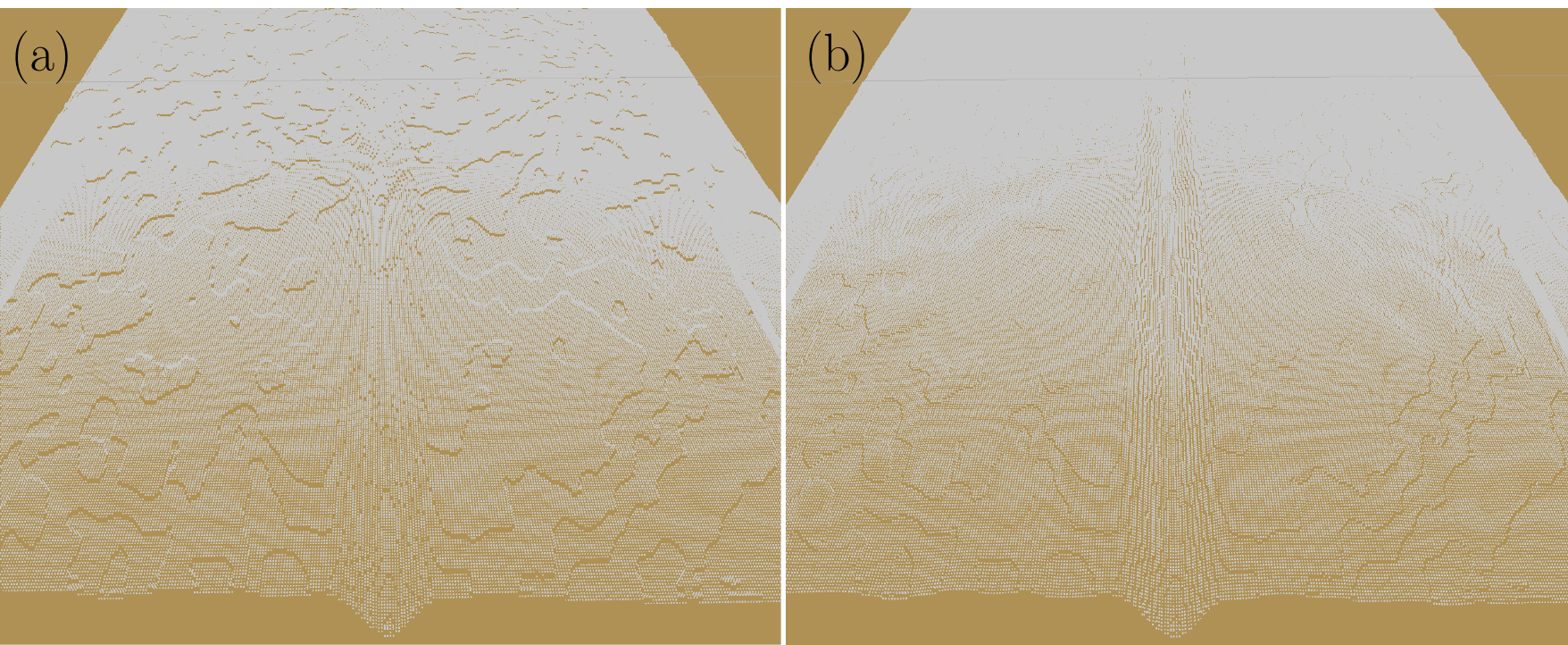}
\caption{Raw point cloud smoothing. 
        (a) Point cloud before smoothing has too many noisy regions.
        (b) Point cloud after smoothing has more of an even surface.}
\label{fig:smooth}
\end{figure}

In general, there is noise data in the raw point cloud owing to camera hardware factors. 
Thus, the point cloud needs smoothing before proceeding to the next step.
PCL provides a Moving Least Squares (MLS) surface reconstruction method to smooth the point cloud surface and reduce noisy data. 
Fig. \ref{fig:smooth} shows the before and after results of the smoothing process for the point cloud of testing the welding workpieces in Fig. \ref{fig:captures}. 
To some extent, it demonstrates that the point cloud surface becomes more even and gentle after smoothing.

Surface normal computation is an important fundamental part of welding groove detection. 
Also, PCL offers a mathematical method to estimate the surface normal of each point. Theoretically, given a point cloud cluster, 
computing one point normal is actually a problem of estimating a normal of a plane tangent to the point cloud surface. Of course, the plane must pass through this point. 
Simply put, one point’s neighbor (a sphere with a constant radius and the point being the sphere’s center) is a small cluster of a point cloud which can fit onto a plane.
Therefore, normal of this plane is regarded as normal of the point.

\begin{figure}[htbp]
\centering
\includegraphics[width=\columnwidth, height=4cm]{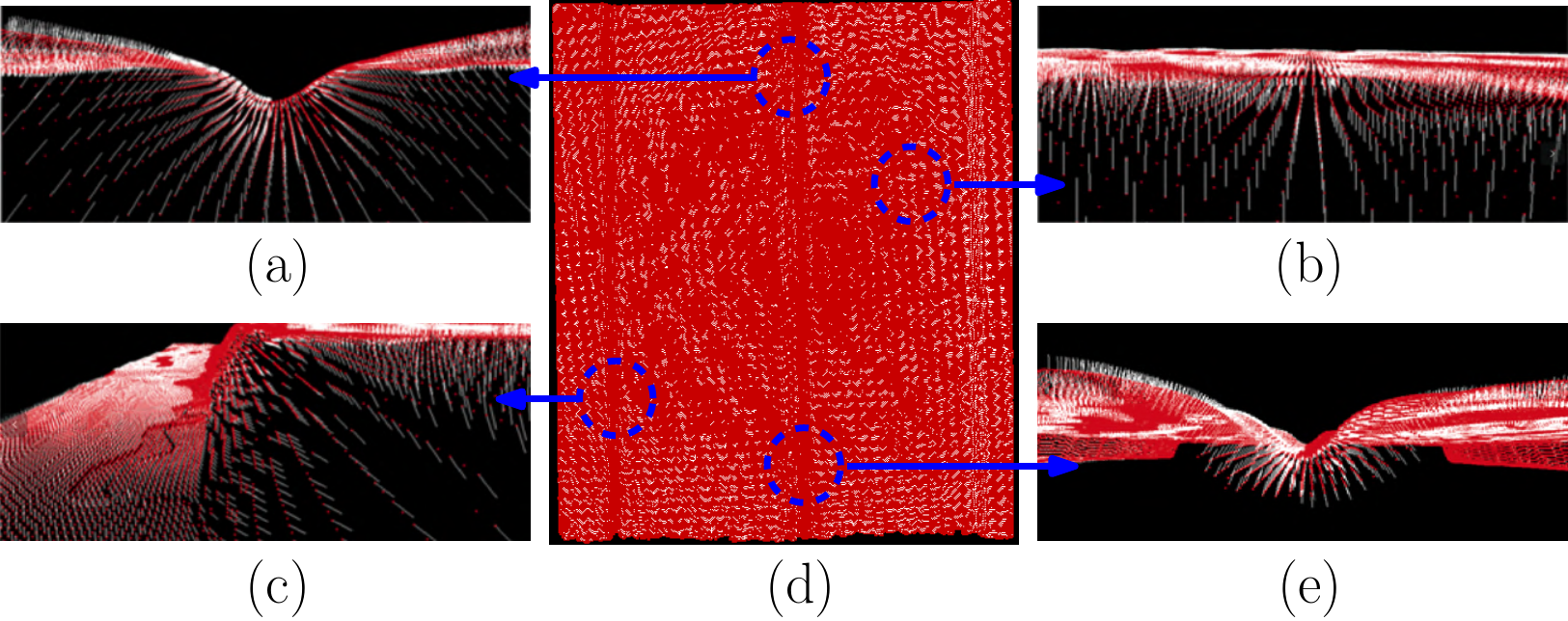}
\caption{Surface normal map. 
            Each white arrow represents the normal of each point. 
            (a) Groove region. 
            (b) Flat region. 
            (c) Edge region. 
            (d) Surface point cloud. 
            (e) Groove region.}
\label{fig:normalMap}
\end{figure}

Then, a normalized map of the surface normal (shown in Fig. \ref{fig:normalMap}) for the after-smoothing point cloud is obtained by iterating every point with least-square plane fitting. 
Moreover, each point normal $\Vec{u_i}$ shown as a white arrow in Fig. \ref{fig:normalMap} is a unit vector defined as:

\begin{equation}
 \Vec{u_i} = [x_i, y_i, z_i]^T, \quad \sqrt{ (x_i)^2 + (y_i)^2 + (z_i)^2 } = 1\label{unit-vetor}
\end{equation}
where $i$ represents the $i$-th point in the organized point cloud. 

Inspired by \cite{rusu2010fast} regarding object recognition,
a descriptive method called a groove feature histogram (GFH) was designed to quantify surface variation extent. 
There are two types of GFH considered for each point in a point cloud: local GFH and global GFH.

\subsection{Local GFH}\label{subsec:Local-GFH}

\begin{figure}[htbp]
\centering
\includegraphics[width=\columnwidth, height=4cm]{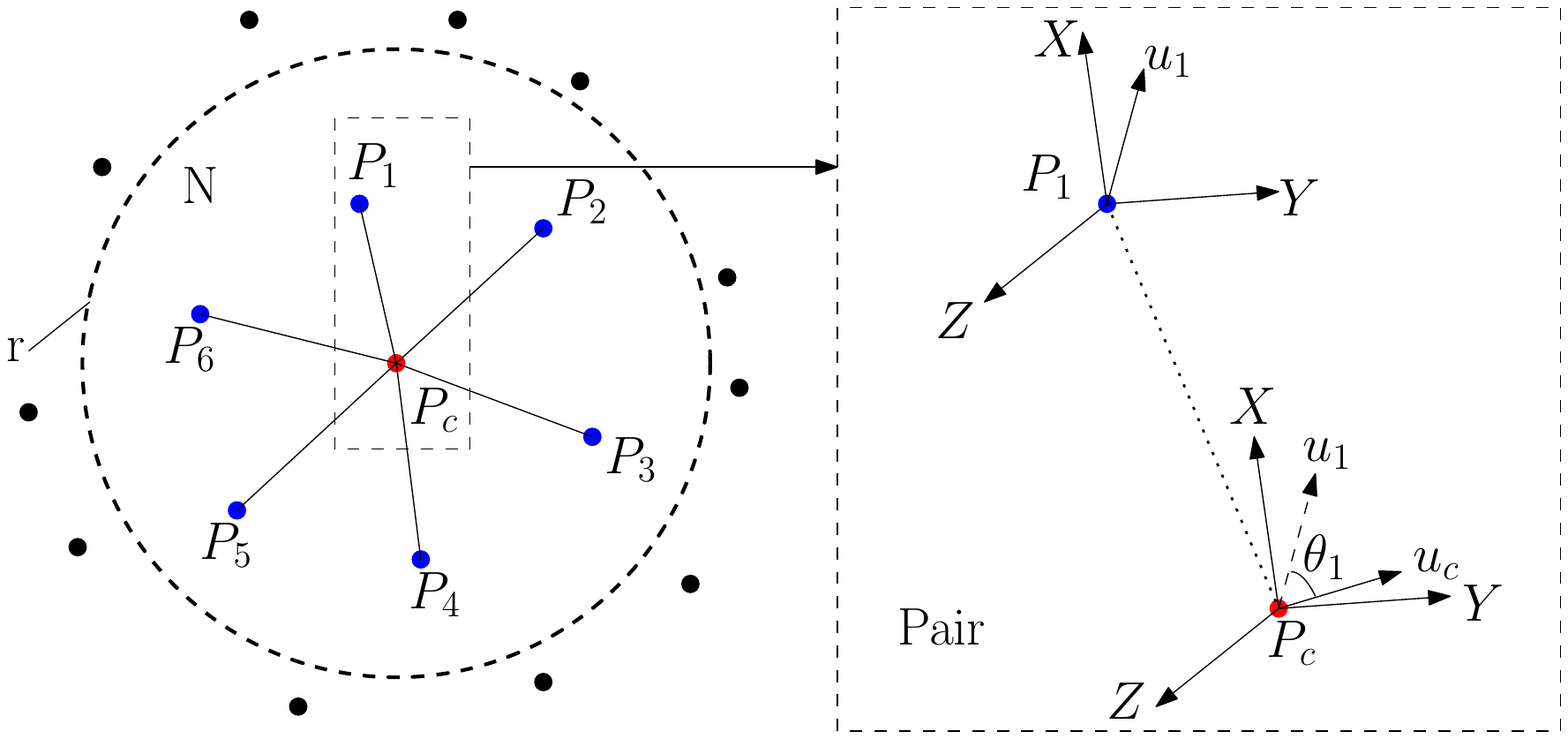}
\caption{(Left) The neighbor $N$ of one point $P$ is defined as a sphere whose radius is $r$.
          Within the neighbor $N$, 
          the red point is the central point $P_{c}$ of the sphere and the blue points are neighbor points.
          (Right) The central point $P_{c}$ with its normal $u_{c}$ is paired to one neighbor point $P_{1}$ with its normal $u_{1}$. 
          $\theta_{1}$ is the included angle between $u_{c}$ and $u_{1}$.}
\label{fig:localGFH}
\end{figure}

The neighbor of each point is defined as a sphere (the point is its neighbor’s central point) with a constant radius $r$.
Within this neighbor, the kdTree search method is used to find other neighbor points (the Euclidean distance to the central point is not more than $r$).
Neighbor points are regarded as an individual point set where the normal of each point can be obtained from the surface normal map (Fig. \ref{fig:normalMap}).
Then the central point is paired with every other neighbor point and every pair has two unit vectors which form one included angle. 
Fig. \ref{fig:localGFH} shows one point's neighbor with one pair.

\begin{figure}[htbp]
\centering
\includegraphics[width=\columnwidth, height=4cm]{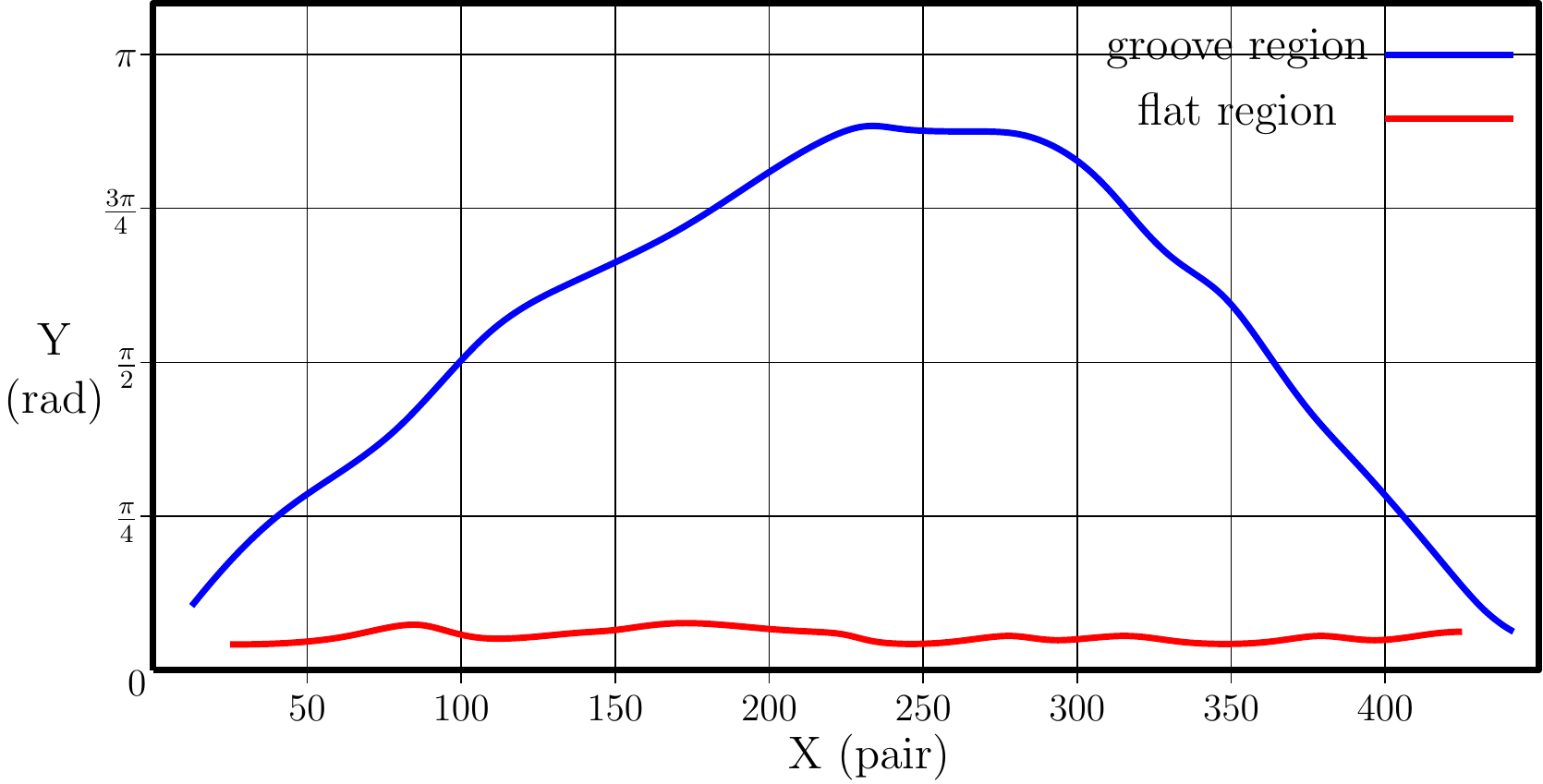}
\caption{Local groove feature histogram. 
         The groove region has higher variation extent than the flat region.
         X-axis shows the number of pairs in one point’s neighbor. 
         Y-axis shows the included angle of one pair.}
\label{fig:localGFH_histogram}
\end{figure}

Generally, since the 3D position of every point is relevant to the manipulator base coordinate system through rigid transformation, 
all the point normals are based on the same coordinate $XYZ$ (see Fig. \ref{fig:localGFH} [Right]).
Furthermore, the included angle $\theta_{j}$ of the $j$-th pair within the center point's neighbor is computed as:
 
\begin{equation}
\theta_{j} = \arccos{ \frac{ \Vec{u_c} \cdot \Vec{u_{j}} }{ \left|\Vec{u_c}\right| \cdot \left|\Vec{u_{j}}\right| } } \label{included-angle}
\end{equation}

where $\Vec{u_c}$ is normal of the center point and 
$\Vec{u_{j}}$ is normal of the $j$-th neighbor point paired to the center point. 
By using Equation (\ref{included-angle}) for iterating every pair of the neighbor,  
all the computation results are arranged as a set ($\Theta = \{\theta_{1}, \theta_{2}, ..., \theta_{n}\}$, $n$ is number of the neighbor points),
which builds up the local GFH of the center point.
Fig. \ref{fig:localGFH_histogram} shows the local GFH of one point from the groove region and one point from the flat region.

\subsection{Global GFH}\label{subsec:Global-GFH}

For global GFH, the whole point cloud needs to be taken into account. 
Therefore, the unit benchmark normal $\Vec{u_b}$ representing the main direction of the whole point cloud is defined as: 
\begin{equation}
\Vec{U_b} = \sum_{i=0}^{n}\Vec{u_i}, \quad \Vec{u_b} = \frac{\Vec{U_b}}{\left|\Vec{U_b}\right|}\label{benchmark2}
\end{equation}
where $\Vec{u_i}$ is the $i$-th point normal of the point cloud.

\begin{figure}[htbp]
\centering
\includegraphics[width=\columnwidth, height=6.5cm]{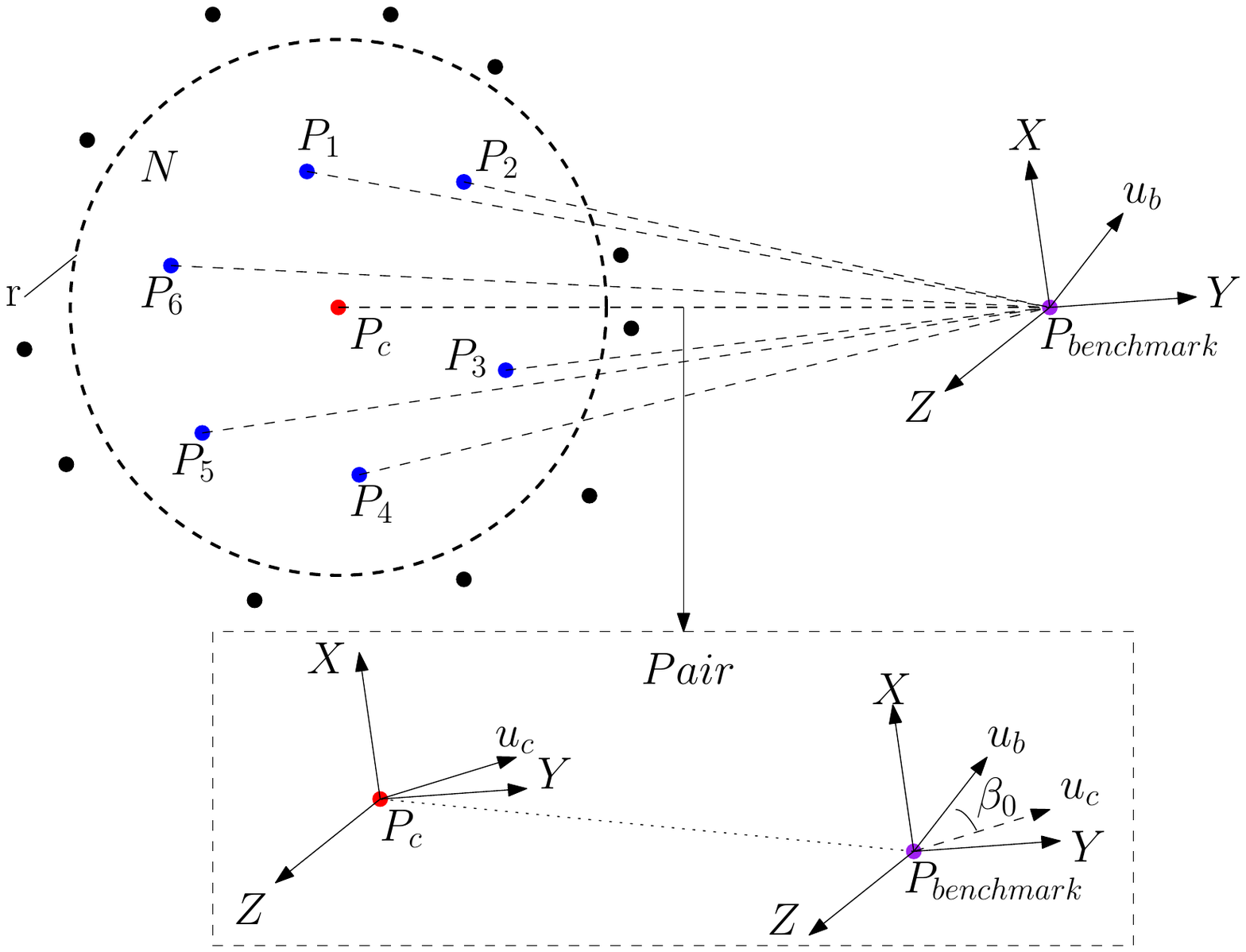}
\caption{Every point normal of the neighbor is paired to the benchmark normal.
         For example of one pair (the central point normal $\Vec{u_c}$ and the benchmark normal $\Vec{u_b}$), 
         $\beta_0$ is the included angle between $\Vec{u_c}$ and $\Vec{u_b}$.}
\label{fig:globalGFH}
\end{figure}

Back to the neighbor of the previous point (the central point $P_c$ in Fig. \ref{fig:localGFH}), 
each point normal in the neighbor is paired to the benchmark normal $\Vec{u_b}$ (Fig. \ref{fig:globalGFH}).

\begin{figure}[htbp]
\centering
\includegraphics[width=\columnwidth, height=4cm]{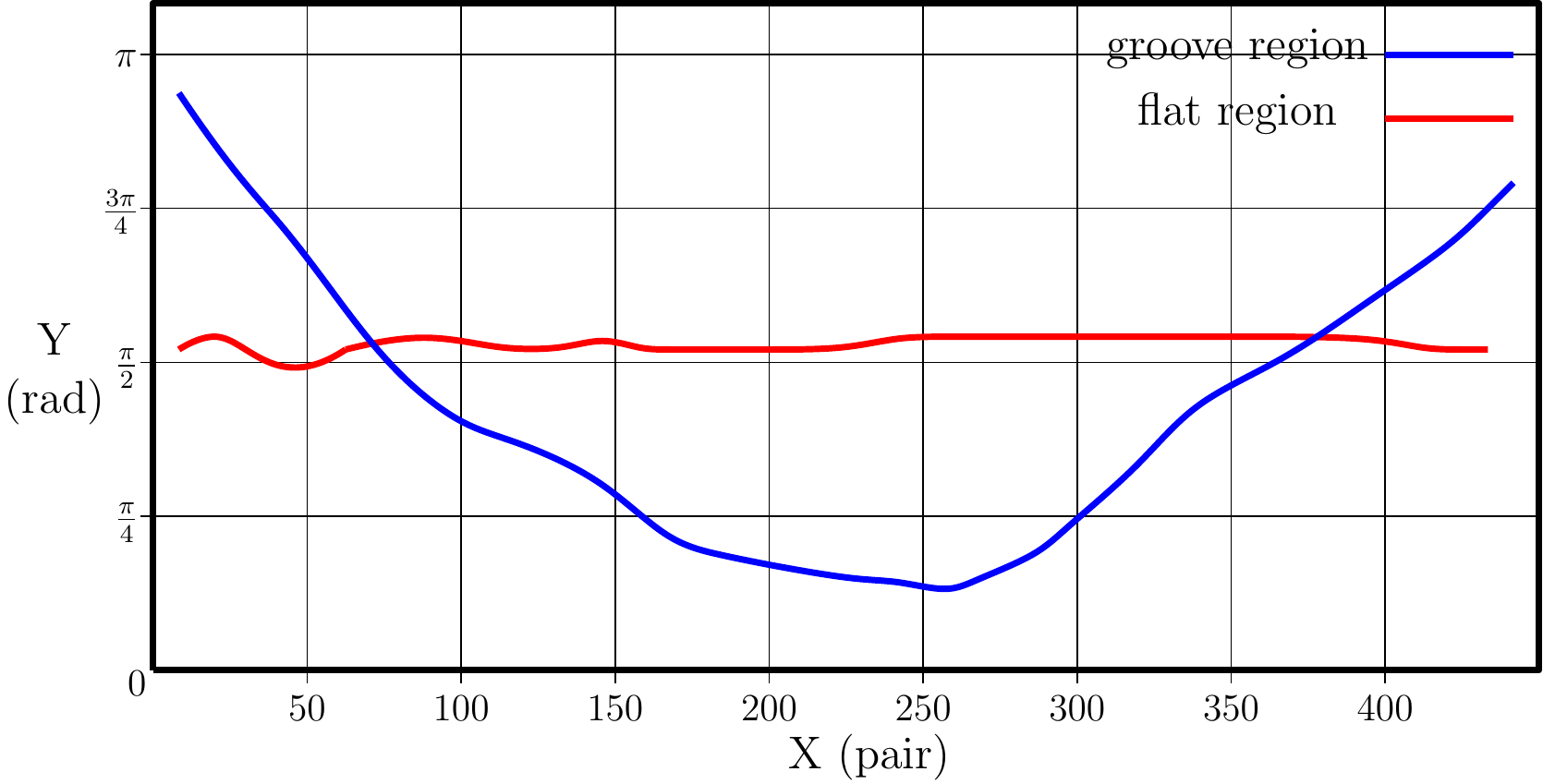}
\caption{Global groove feature histogram. 
        The groove region has a higher variation extent than the flat region.
        X-axis shows the number of pairs in one point’s neighbor. 
         Y-axis shows the included angle of one pair.}
\label{fig:globalGFH_histogram}
\end{figure}

Under the same principle as local GFH, the $k$-th included angle $\beta_{k}$ of the neighbor is computed as:
\begin{equation}
\beta_{k} = \arccos{ \frac{ \Vec{u_b} \cdot \Vec{u_{k}} }{ \left|\Vec{u_b}\right| \cdot \left|\Vec{u_{k}}\right| } } \label{globalGFH}
\end{equation}

Therefore all the resultant included angles are put into one set ($B = \{\beta_{1}, \beta_{2}, ..., \beta_{n}\}$) which
builds up the global GFH of the central point.
Fig. \ref{fig:globalGFH_histogram} shows the global GFH of one point from the groove region and one point from the flat region.

\subsection{Surface Variation Descriptor}

According to analysis of the local GFH \ref{subsec:Local-GFH} and global GFH \ref{subsec:Global-GFH},
the variations for both histograms of one point are defined respectively as:
\begin{equation}
    \sigma^l = \frac{\sum_{j=0}^{n}{ (\theta_j - \overline{\theta})^2 } }{n}, \quad \sigma^g = \frac{\sum_{k=0}^{n}{ (\beta_k - \overline{\beta})^2 } }{n}\label{variation}
\end{equation}
where $\overline{\theta}$ is average value of the local GFH and $\overline{\beta}$ is average value of the global GFH. 
$n$ is the number of pairs in the point's neighbor.
Then the surface variation descriptor $D_i$ for $i$-th point $P_i$ of the point cloud is defined as:
\begin{equation}
D_i = \sqrt{(\sigma_i^l)^2 + (\sigma_i^g)^2} \label{globalGFH}
\end{equation}
where $\sigma_i^l$ is variation of local GFH of $P_i$ and $\sigma_i^g$ is variation of global GFH of $P_i$. 

\begin{figure}[htbp]
\centering
\includegraphics[width=\columnwidth, height=3cm]{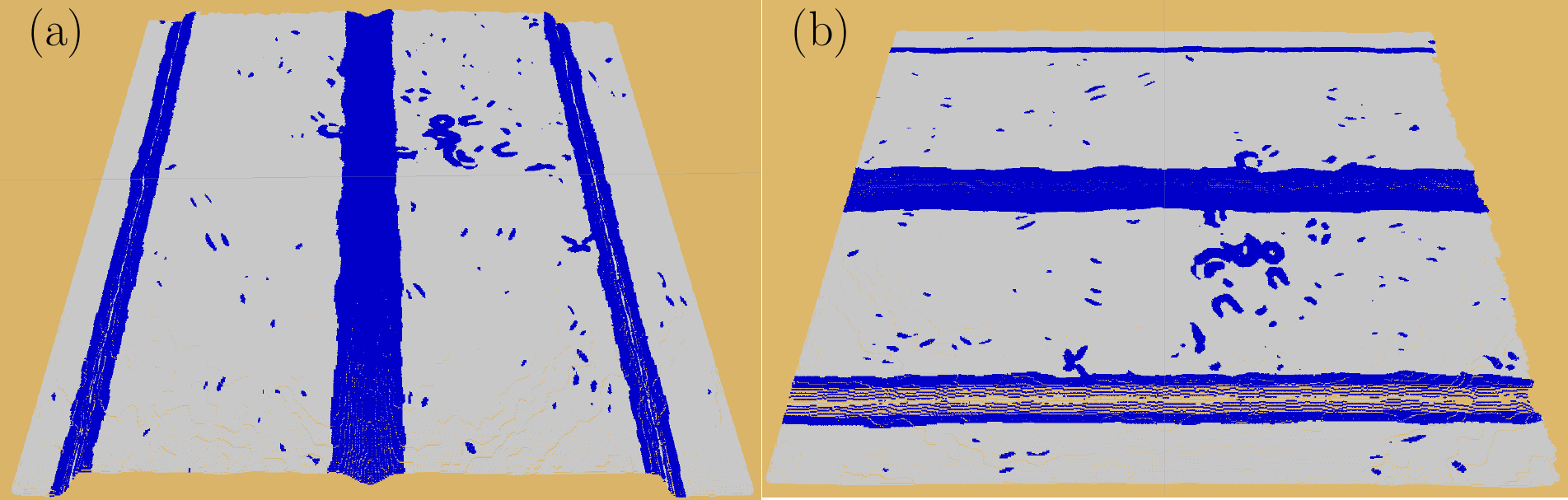}
\caption{Surface Variation Map. (a) Front view. (b) Side view.}
\label{fig:result}
\end{figure}

Through computing $D_i$ of every point, a map which represents the surface variation extent of the whole point cloud is obtained and shown as Fig. \ref{fig:result}. 
It intuitively shows the extent of surface variation on the whole point cloud, although there are noisy places.
The blue regions, such as the groove region and the edges, have high variation, while the white regions are almost flat.

\begin{figure}[htbp]
\centering
\includegraphics[width=\columnwidth, height=3cm]{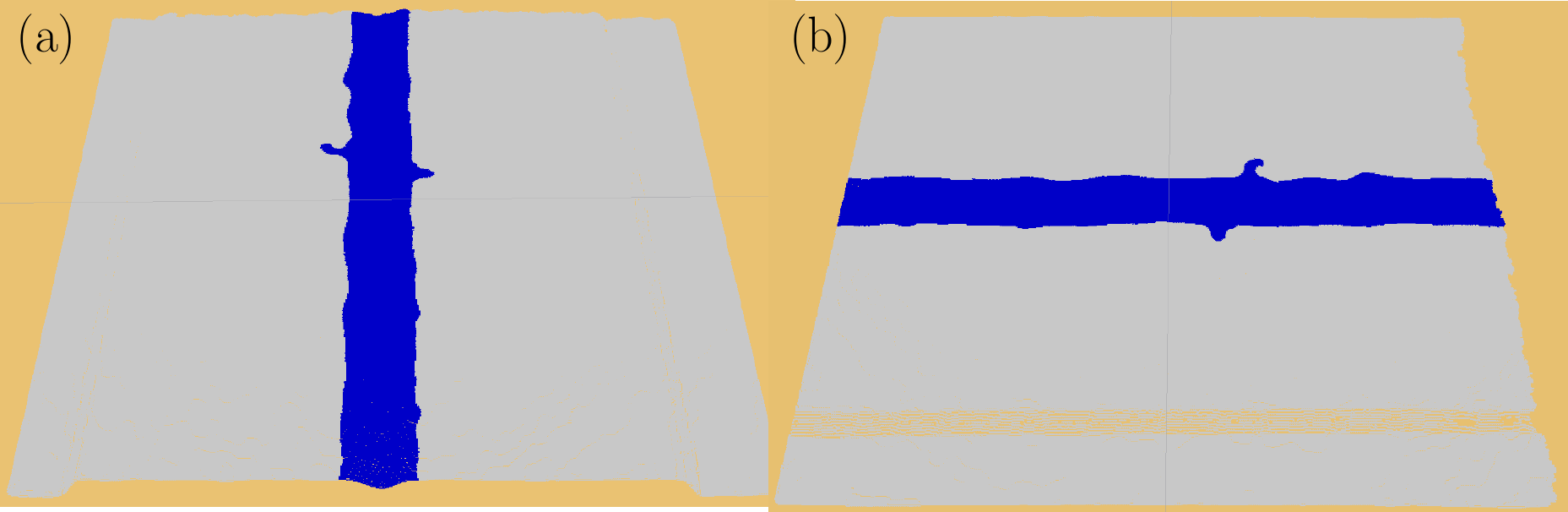}
\caption{Groove Detection Result. (a) Front view. (b) Side view}
\label{fig:Grooveresult}
\end{figure}

By analyzing the surface variation map, 
points of the groove region with descriptor values more than threshold: 4.5-5 are centralized and almost tightly connected. 
In terms of the descriptor threshold, 
all the unqualified points are deleted. 
Consequently, only the groove point set remains. 
The groove detection results (blue region) are shown in Fig. \ref{fig:Grooveresult}. 
In the next step, the groove point set is used to generate the 3D welding trajectory for the robotic arm.

Compared with the  PFH descriptor (computational complexity is $\mathcal{O}( {\nu} {\mu}^2)$) \cite{alexandre20123d}, 
the surface variation descriptor solely considers angular values instead of combining angular values and the distance of a single pair. 
One of the two components of the surface variation descriptor, global GFH, is related to all the point normals, 
meaning that it can adapt to different types of workpieces.
Since PFH focuses on local geometric features, 
it is not as adaptive as global GFH. 
Moreover, the other component, local GFH, is simplified SFPH \cite{rusu2009fast} without involving the distance information. 
Therefore, the proposed surface variation descriptor reduces the computational complexity to $\mathcal{O}(2 {\nu} {\mu})$, 
where the ${\nu}$ is the number of points in the point cloud and ${\mu}$ is the number of pairs in each point's neighbor.
In an actual experiment, the proposed algorithm could run in real-time.

\section{WELDING TRAJECTORY GENERATION}\label{sec:path-generation}

\begin{figure}[htbp]
\centering
\includegraphics[width=\columnwidth, height=3.5cm]{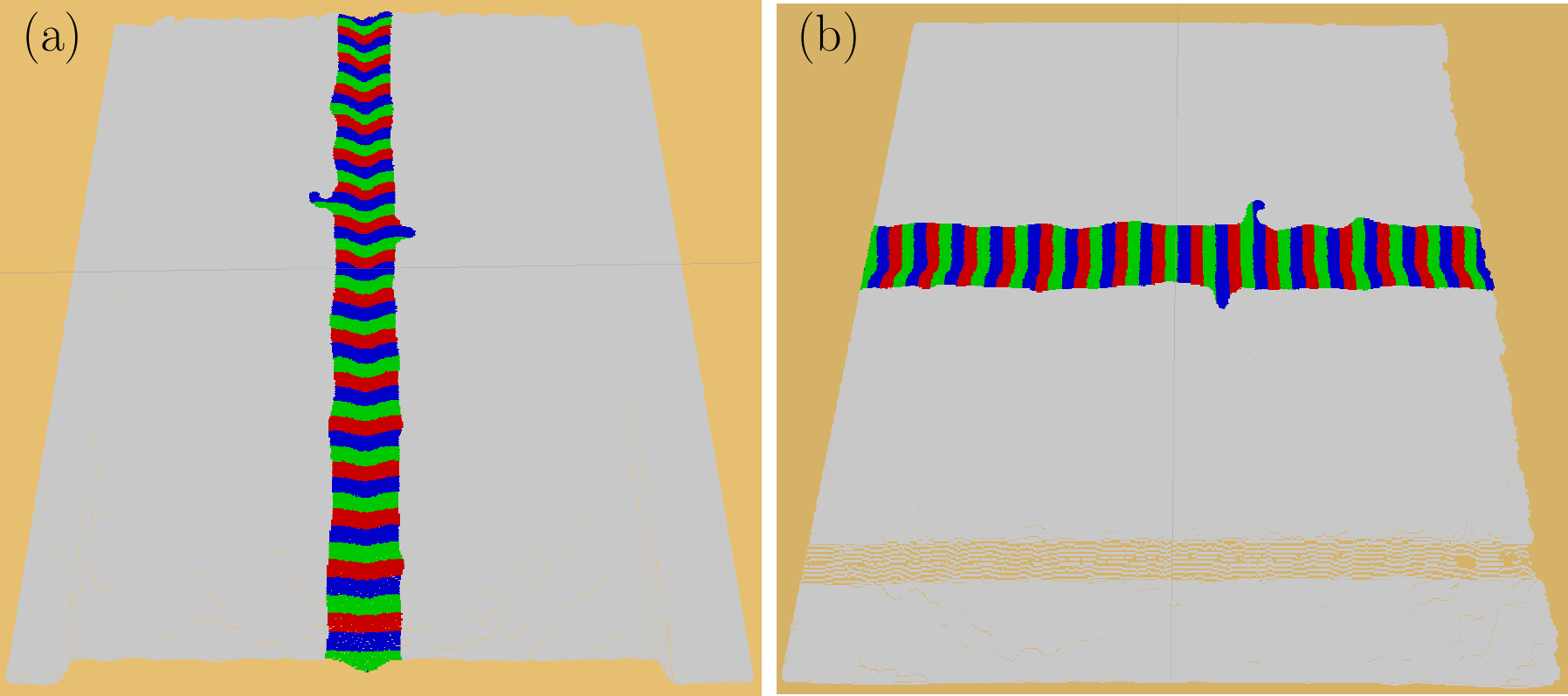}
\caption{Groove point set segmentation. (a) Front view of segmented groove region. (b) Side view of segmented groove region.}
\label{fig:Segmentation}
\end{figure}

Based on the groove point set (marked in blue) obtained by the groove detection algorithm (see Fig. \ref{fig:Grooveresult}), 
the 3D welding trajectory is generated accordingly.
The welding motion direction is closely dependent on the layout of the groove point set.
Then, along the direction, the point set is evenly segmented into 50-60 consecutive regions of the same width (blue, green, and red for distinction), as shown in Fig. \ref{fig:Segmentation}.
The width of each segmented region is related to the total length of the welding groove; 
and each segmented region generates a single way point. 
Together, all the way points form the final 3D welding trajectory.

The sum of the distance from the way point of each segmented region to every point of the same segmented region should be as short as possible. 
The generation accuracy of the way points is dependent on groove detection results. 
If one resultant segmented region does not perfectly match the actual groove, 
its way point will deviate from the central position. 
The issue can be defined as a function for each segmented region:

\begin{equation}
\Vec{I_i^s} = P^w - P_i^s, \quad f = argmin \sum_{i=0}^m \left| \Vec{I_i^s} \right| \label{argmin}
\end{equation}
where $P^w$ is the unknown way point and $P_i^s$ is the $i$-th point of one segmented region. 
When $f$ gets to its minimum, $P^w$ is obtained as the way point.
During experimentation, 
the gradient descent method is used to optimize the function. 
Specifically, the value 1e-4 is set as the threshold for the aborting iteration and the maximum of the iteration loop is set as 1000.

\begin{figure}[htbp]
\centering
\includegraphics[width=\columnwidth, height=3.7cm]{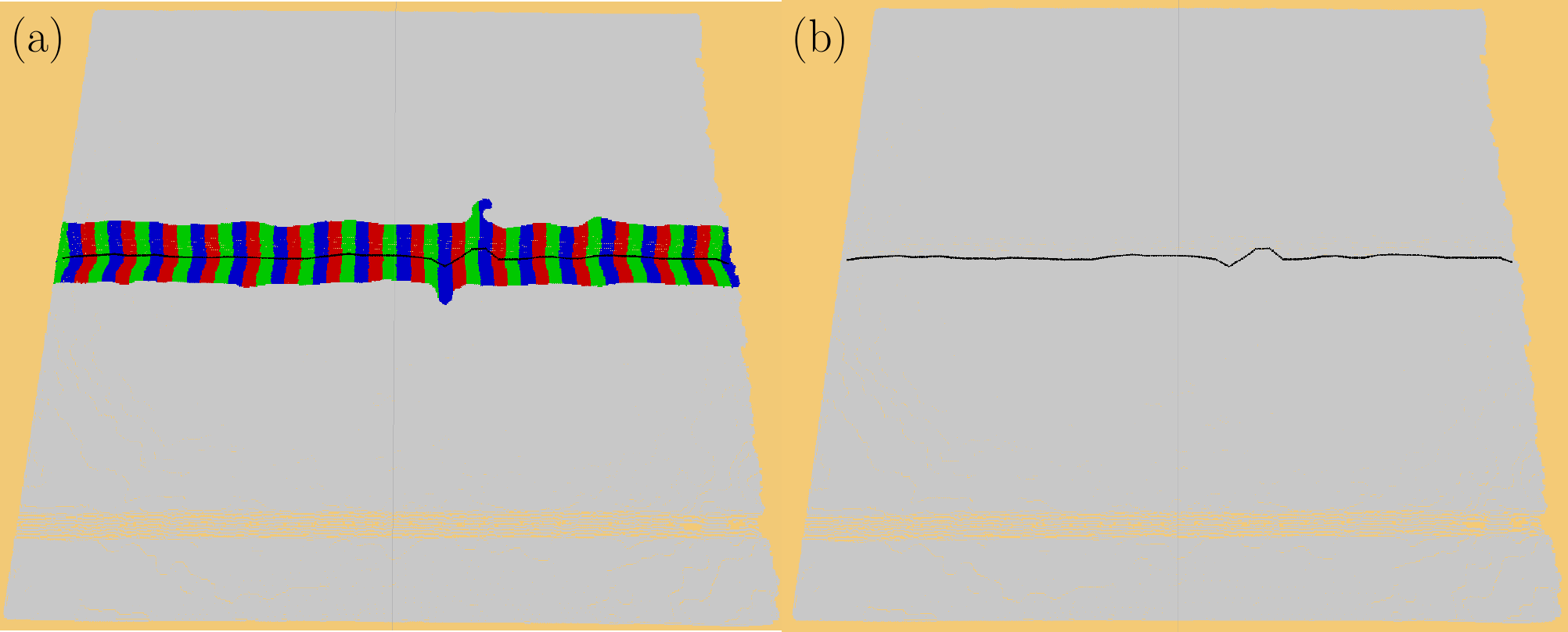}
\caption{Welding trajectory generation. 
         (a) Generated welding trajectory inside the segmented groove point set. 
         (b) The trajectory inside the raw point cloud.}
\label{fig:trajectory-generation}
\end{figure}

After iterating every segmented region of the groove point using Equation (\ref{argmin}), 
the final welding trajectory generation result is shown as Fig. \ref{fig:trajectory-generation}.
The 3D orientation of each way point of the generated trajectory is represented as a unit vector $\Vec{o_w}$, defined as:
\begin{equation}
\Vec{O_w} = \sum_{i=0}^m \Vec{u_i^s}, \quad \Vec{o_w} = \frac{\Vec{O_w}}{\left| \Vec{O_w} \right|}\label{orientation}
\end{equation}
where $\Vec{u_i^s}$ is the $i$-th point normal of the segmented region. 
Then the unit vector is transformed as three Euler angles by the Eigen library.
Ultimately, a 3D welding trajectory with both 3D positions and orientations for the welding groove in workpiece is generated.
The trajectory will then be sent to the manipulator control for executing the actual tracking motion.

\section{RESULTS}\label{sec:results}

\begin{figure}[htbp]
\centering
\includegraphics[width=\columnwidth, height=3cm]{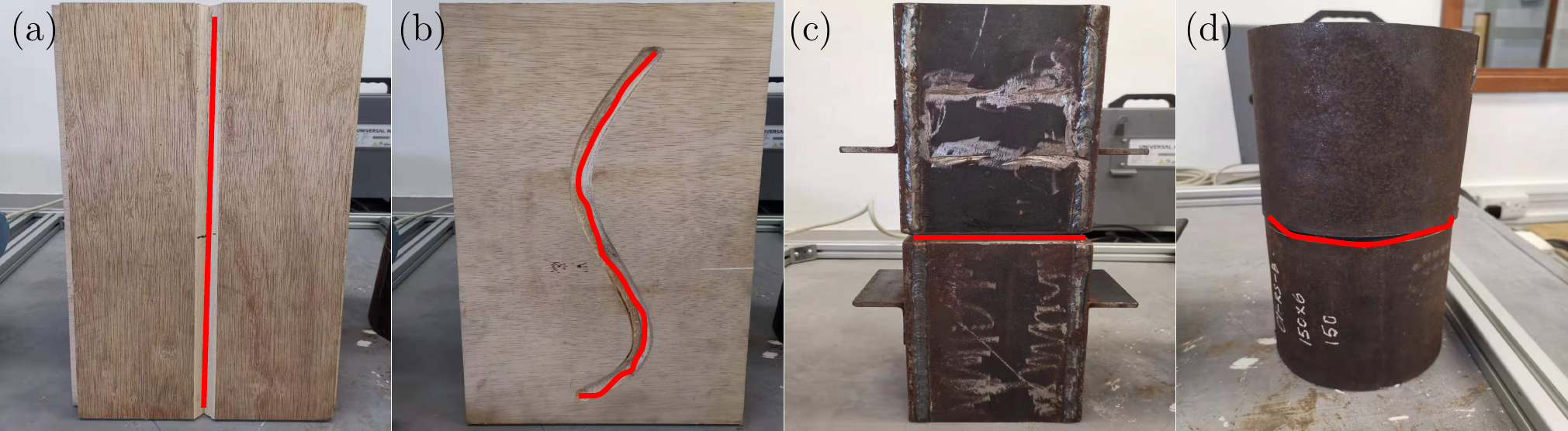}
\caption{Experimental welding workpieces with their welding groove marked as red regions. 
        According to the shape of the welding groove, 
        the workpiece is defined as: 
        (a) straight-line; 
        (b) curve-line; 
        (c) box; 
        (d) cylinder.}
\label{fig:workpiece_groove}
\end{figure}

An open welding environment (Fig. \ref{fig:setup}) with four types of welding workpieces was prepared for experiments. 
Fig. \ref{fig:workpiece_groove} shows the experimental welding workpieces which are straight-line, curve-line, box and cylinder respectively.
The reason why the wood workpieces are regarded as experimental objects, is due to their convenience to
simulate real welding groove like straight type or curve type. The material of welding workpieces has no effects on the proposed algorithm, 
because the inputting 3D point cloud contains no optical properties.

In order to objectively evaluate the performance of the proposed method, 
three vital elements are considered:
\begin{enumerate}
    \item The whole processing runtime from the raw point cloud to generating the motion trajectory.
    \item The overlapping rate of the detected groove region and the actual groove region.
    \item The disparity between the generated trajectory and standard welding trajectory.
\end{enumerate}
The processing runtime as an essential factor could evaluate the efficiency of the proposed robotic system for automatic welding. 
The overlapping rate is used to illustrate the accuracy of the welding groove detection algorithm.
The disparity is presented by actual automatic welding execution.
Thus, by focusing on the aforementioned three elements, each workpiece shown in Fig. \ref{fig:workpiece_groove} was tested.
In fact, although the RGB-D camera is well calibrated, an error in the measurement of depth still exists.

\subsection{Processing Runtime Results}

To measure the processing runtime of the proposed method for the surface point cloud of one workpiece, 
the function “clock()” of C++ is used to capture the system start time and end time of the method. Due to the running performance of the PC, 
each workpiece was tested ten times and each runtime from inputting the raw point cloud to generating the motion trajectory was recorded (Table \ref{tab:Runtime}).

\begin{table}[htbp]
    \begin{center}
    \caption{Results of processing runtime for each workpiece} \label{tab:Runtime}
        \begin{tabular}{lcc}
            \toprule
            Types &$P_{num}$   &$t_{mean}$(s) \\ 
            
            \midrule
            Straight-line &265800    &14.09\\
            
            Curve-line    &266497     &14.08\\
            
            Box           &127031  &7.51\\ 
            
            Cylinder      &121429 &6.05\\
            

             
            
             
             
            \bottomrule
        \end{tabular}
 
    \vspace{5pt}
    \text{$P_{num}$ = Number of points. $t_{mean}$ = Average time of runtime.}
    \end{center}
\end{table}

\subsection{Groove Detection Results}

To evaluate the accuracy of the welding groove detection algorithm, the results of the algorithm need to be compared with the ground truth (actual groove region defined by the authors). 
Inspired by the concept of IoU (intersection over union) in 2D image processing, the 3D overlapping rate $\lambda$ of the detected groove region and actual groove region is introduced:
\begin{equation}
    \lambda = \frac{n_{overlap}}{n_{d} + n_{gt} - n_{overlap}}\label{overlapping-rate}
\end{equation}
where $n_{d}$ is the number of points of the detected groove region, $n_{gt}$ is the number of points of the actual groove region and $n_{overlap}$ is number of points of overlapping region between the detected groove region and the actual groove region.

\begin{figure}[htbp]
\centering
\includegraphics[width=\columnwidth, height=8cm]{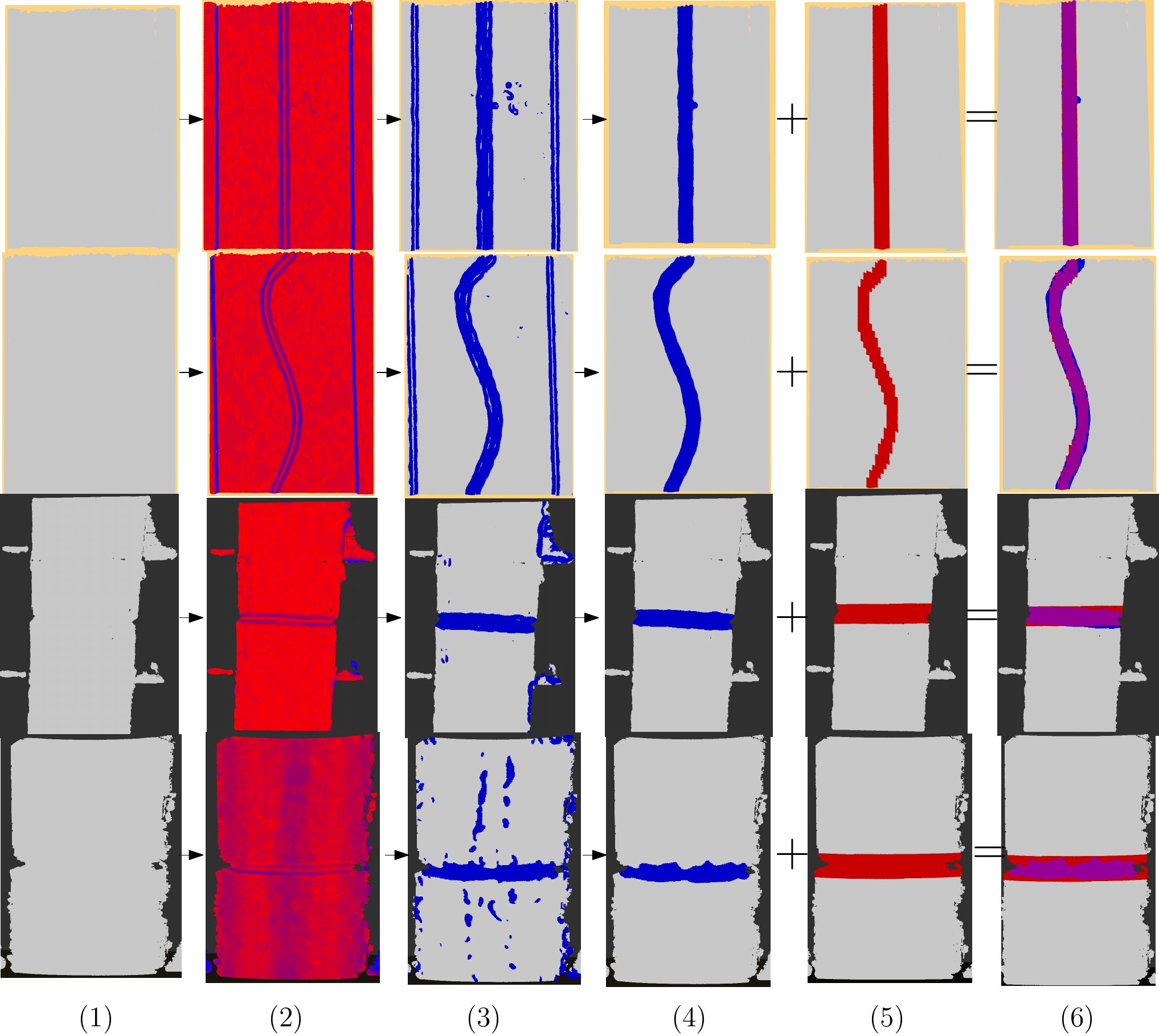}
\caption{Welding groove detection workflow. 
        (1) Input raw point cloud of each workpiece. 
        (2) Global normal map. 
        (3) Surface variation map by groove feature histogram (GFH). 
        (4) Filtering results (detected groove region). 
        (5) Ground truth of groove region in raw point cloud. 
        (6) Overlapping map by adding detected groove region and ground truth of groove region. 
        Eventually the overlapping map is used to compute the 3D overlapping rate to evaluate the accuracy of the groove detection algorithm.}
\label{fig:detectionResult}
\end{figure}

The welding groove detection process for each workpiece in Fig. \ref{fig:workpiece_groove} is shown in Fig. \ref{fig:detectionResult}, and the detection accuracy (defined by 3D overlapping rate) results are presented in Table \ref{tab:detection}. 

\begin{table}[htbp]
    \begin{center}
    \caption{Results of groove detection accuracy for each workpiece} \label{tab:detection}
    \resizebox{\columnwidth}{!}{
        \begin{tabular}{lcccccc}
            \toprule
            Types &$\lambda_1$ &$\lambda_2$  &$\lambda_3$ &$\lambda_4$ &$\lambda_5$ &$\lambda_{mean}$\\ 
            
            \midrule
            Straight-line &92.74 &93.01 &92.81 &92.49 &91.35 &92.48\\     
            Curve-line    &81.24 &82.58 &82.13 &81.87 &82.29 &82.02\\      
            Box           &81.48 &82.65 &82.02 &80.49 &81.97 &81.72\\      
            Cylinder      &63.27 &61.79 &64.69 &67.57 &65.75 &64.61\\      
            \bottomrule
        \end{tabular}
 
    }
    \vspace{2pt}
    \newline
    \text{$\lambda_1$-$\lambda_5$ = Accuracy (\%) of each test. $\lambda_{mean}$ = Average accuracy.}
    \end{center}
\end{table}

For types of straight-line, curve-line and box, the proposed welding groove detection approach
has quite high detection accuracy. But for type of cylinder, the performance of the detection approach is not acceptable,
because of poor robustness to model surface which is not almost plane.

\subsection{Motion Execution Results}

\begin{figure}[htbp]
\centering
\includegraphics[width=\columnwidth, height=8cm]{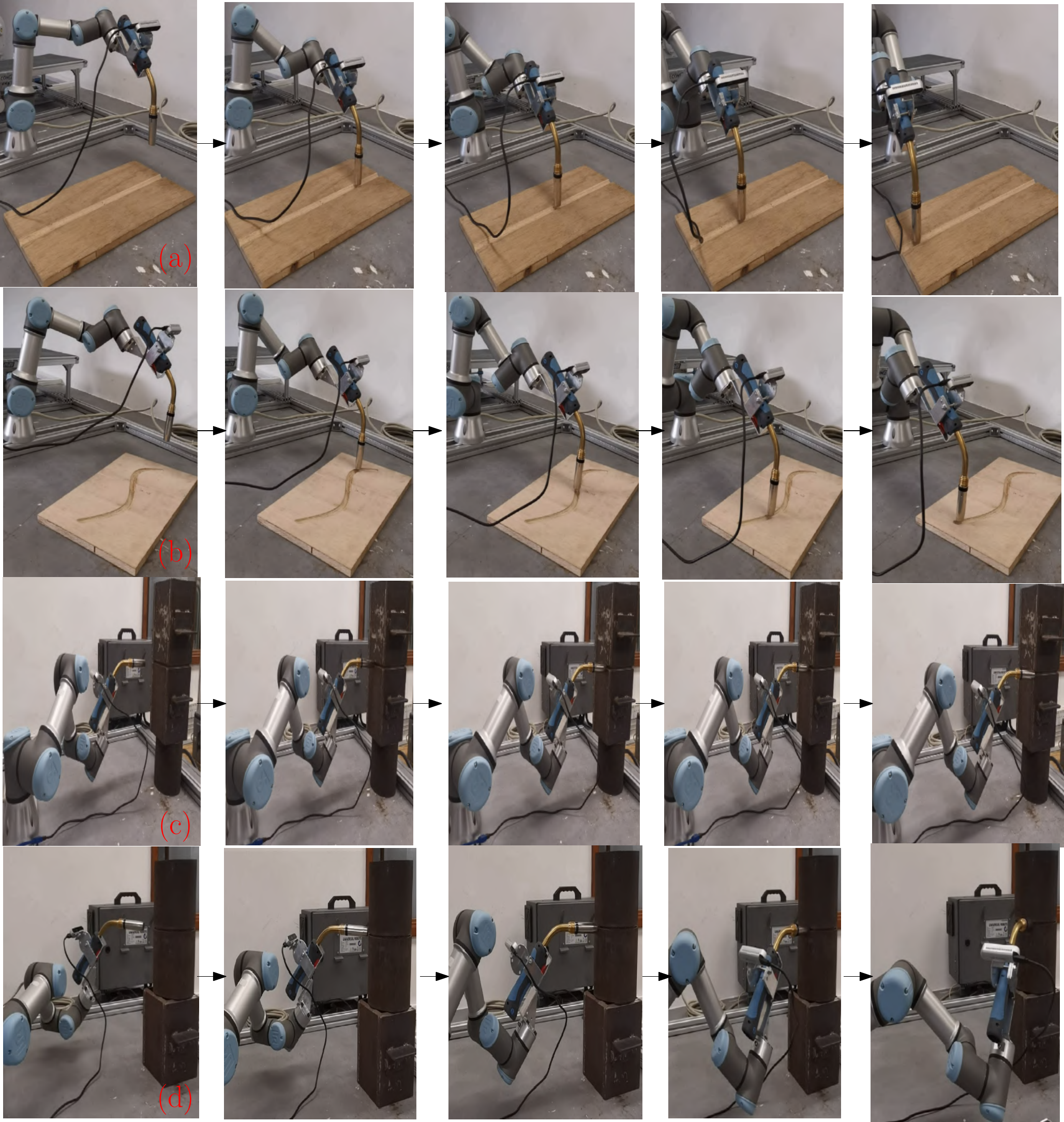}
\caption{Actual motion execution of the robotic manipulator. (a)-(b) represents type of straight-line, curve-line, metal box and metal cylinder workpiece respectively. And their motion order is from left side to right side.}
\label{fig:motion}
\end{figure}

As discussed in Section \ref{sec:path-generation}, the motion trajectory generation is based on the point set of the detected groove region (see Fig. \ref{fig:detectionResult}).
Therefore, the robotic manipulator drives the welding torch which follows the motion trajectory to execute the welding tasks.
Fig. \ref{fig:motion} presents the actual motion execution of the robotic manipulator.
As shown in Fig. \ref{fig:motion}, the motion of the manipulator fits the generated trajectory tightly without terrible mismatch.
Also, by using cartesian path planning of the Moveit \cite{chitta2012moveit} package, the welding torch is able to complete welding tasks well (smoothly following the trajectory of the groove).
However, it is difficult to evaluate the welding torch's motion accuracy through external devices such as Vicon \cite{clarisse1986vicon}. 
In this case, evaluating the actual welding quality is the most reasonable course of action.

According to the running performance (runtime and detection accuracy) of the proposed method, 
the robotic system has the capability to realize automatic welding tasks efficiently, 
as compared to the conventional teach-playback method. 

\section{Conclusions}\label{sec:conclusions}
 
This paper presents an integrated robotic system for industrial welding with an automatic groove detection and trajectory generation method. 
The system is composed of a robotic manipulator (Universal Robot 3),
an RGB-D camera (Realsense D415), and a welding torch. 
The system also has good flexibility when facing different welding situations. 
The software framework is built on ROS, with 3D point cloud processing being key. 
Four types of general welding workpieces were tested. 
After evaluating the accuracy between the welding trajectory generated by the proposed method and the ground truth of the welding groove, 
the motion execution performance proves good feasibility of the designed robotic system. 
However, a problem with the proposed method is that it cannot cope with a larger 3D point cloud of welding workpiece surfaces due to the more complex geometric regions and noise. 
In future work, a neural network instead of a geometric feature-based method could be introduced to improve robustness and accuracy of the welding groove detection algorithm. 
Also, we may incorporate additional functionalities to the robot, e.g. the capability to interact with the workpice \cite{dna2014_tcst}.

\bibliographystyle{IEEEtran}
\bibliography{rick_biblio.bib}

\begin{thebibliography}{10}
\providecommand{\url}[1]{#1}
\csname url@rmstyle\endcsname
\providecommand{\newblock}{\relax}
\providecommand{\bibinfo}[2]{#2}
\providecommand\BIBentrySTDinterwordspacing{\spaceskip=0pt\relax}
\providecommand\BIBentryALTinterwordstretchfactor{4}
\providecommand\BIBentryALTinterwordspacing{\spaceskip=\fontdimen2\font plus
\BIBentryALTinterwordstretchfactor\fontdimen3\font minus
  \fontdimen4\font\relax}
\providecommand\BIBforeignlanguage[2]{{%
\expandafter\ifx\csname l@#1\endcsname\relax
\typeout{** WARNING: IEEEtran.bst: No hyphenation pattern has been}%
\typeout{** loaded for the language `#1'. Using the pattern for}%
\typeout{** the default language instead.}%
\else
\language=\csname l@#1\endcsname
\fi
#2}}

\bibitem{norberto2004cad}
J.~Norberto~Pires, T.~Godinho, and P.~Ferreira, ``Cad interface for automatic
  robot welding programming,'' \emph{Industrial Robot: An Int. J.}, vol.~31,
  no.~1, pp. 71--76, 2004.

\bibitem{li2009measurement}
Y.~Li, Y.~F. Li, Q.~L. Wang, D.~Xu, and M.~Tan, ``Measurement and defect
  detection of the weld bead based on online vision inspection,'' \emph{IEEE
  Trans. Instrumentation and Measurement}, vol.~59, no.~7, pp. 1841--1849,
  2009.

\bibitem{liu2014iterative}
Y.~Liu and Y.~Zhang, ``Iterative local anfis-based human welder intelligence
  modeling and control in pipe gtaw process: A data-driven approach,''
  \emph{IEEE/ASME Trans. Mechatronics}, vol.~20, no.~3, pp. 1079--1088, 2014.

\bibitem{diao2017passive}
C.~Diao, J.~Ding, S.~Williams, Y.~Zhao, \emph{et~al.}, ``A passive imaging
  system for geometry measurement for the plasma arc welding process,''
  \emph{IEEE Trans. Industrial Electronics}, vol.~64, no.~9, pp. 7201--7209,
  2017.

\bibitem{li2016new}
J.~Li, F.~Jing, and E.~Li, ``A new teaching system for arc welding robots with
  auxiliary path point generation module,'' in \emph{35th Chinese Control Conf.
  (CCC)}, 2016, pp. 6217--6221.

\bibitem{rodriguez2017feasibility}
M.~Rodr{\'\i}guez-Mart{\'\i}n, P.~Rodr{\'\i}guez-Gonz{\'a}lvez,
  D.~Gonzalez-Aguilera, and J.~Fernandez-Hernandez, ``Feasibility study of a
  structured light system applied to welding inspection based on articulated
  coordinate measure machine data,'' \emph{IEEE Sensors J.}, vol.~17, no.~13,
  pp. 4217--4224, 2017.

\bibitem{ahmed2016object}
S.~M. Ahmed, Y.~Z. Tan, G.~H. Lee, C.~M. Chew, and C.~K. Pang, ``Object
  detection and motion planning for automated welding of tubular joints,'' in
  \emph{IEEE/RSJ Int. Conf. Intelligent Robots and Systems (IROS)}, 2016, pp.
  2610--2615.

\bibitem{ma2010robot}
H.~Ma, S.~Wei, Z.~Sheng, T.~Lin, and S.~Chen, ``Robot welding seam tracking
  method based on passive vision for thin plate closed-gap butt welding,''
  \emph{The Int. J. Advanced Manufacturing Technology}, vol.~48, no. 9-12, pp.
  945--953, 2010.

\bibitem{rao2018real}
S.~H. Rao, V.~Kalaichelvi, and R.~Karthikeyan, ``Real-time tracing of a weld
  line using artificial neural networks,'' in \emph{IEEE/ACIS 17th Int. Conf.
  Computer and Information Science (ICIS)}, 2018, pp. 275--280.

\bibitem{nele2013image}
L.~Nele, E.~Sarno, and A.~Keshari, ``An image acquisition system for real-time
  seam tracking,'' \emph{The Int. J. Advanced Manufacturing Technology},
  vol.~69, no. 9-12, pp. 2099--2110, 2013.

\bibitem{rusu20113d}
R.~B. Rusu and S.~Cousins, ``3d is here: Point cloud library (pcl),'' in
  \emph{IEEE Int. Conf. Robotics and Automation}, 2011, pp. 1--4.

\bibitem{xu2017welding}
Y.~Xu, N.~Lv, G.~Fang, S.~Du, W.~Zhao, Z.~Ye, and S.~Chen, ``Welding seam
  tracking in robotic gas metal arc welding,'' \emph{J. Materials Processing
  Technology}, vol. 248, pp. 18--30, 2017.

\bibitem{ahmed2018edge}
S.~M. Ahmed, Y.~Z. Tan, C.~M. Chew, A.~Al~Mamun, and F.~S. Wong, ``Edge and
  corner detection for unorganized 3d point clouds with application to robotic
  welding,'' in \emph{IEEE/RSJ Int. Conf. Intelligent Robots and Systems
  (IROS)}, 2018, pp. 7350--7355.

\bibitem{song2014sliding}
S.~Song and J.~Xiao, ``Sliding shapes for 3d object detection in depth
  images,'' in \emph{European Conf. computer vision}, 2014, pp. 634--651.

\bibitem{yang2018pixor}
B.~Yang, W.~Luo, and R.~Urtasun, ``Pixor: Real-time 3d object detection from
  point clouds,'' in \emph{Proceedings of the IEEE Conf. Computer Vision and
  Pattern Recognition}, 2018, pp. 7652--7660.

\bibitem{maiolino2017flexible}
P.~Maiolino, R.~Woolley, D.~Branson, P.~Benardos, A.~Popov, and S.~Ratchev,
  ``Flexible robot sealant dispensing cell using rgb-d sensor and off-line
  programming,'' \emph{Robotics and Computer-Integrated Manufacturing},
  vol.~48, pp. 188--195, 2017.

\bibitem{jing2016rgb}
L.~Jing, J.~Fengshui, and L.~En, ``Rgb-d sensor-based auto path generation
  method for arc welding robot,'' in \emph{Chinese control and decision Conf.
  (CCDC)}, 2016, pp. 4390--4395.

\bibitem{patil2019extraction}
V.~Patil, I.~Patil, V.~Kalaichelvi, and R.~Karthikeyan, ``Extraction of weld
  seam in 3d point clouds for real time welding using 5 dof robotic arm,'' in
  \emph{5th Int. Conf. Control, Automation and Robotics (ICCAR)}, 2019, pp.
  727--733.

\bibitem{quigley2009ros}
M.~Quigley, K.~Conley, B.~Gerkey, J.~Faust, T.~Foote, J.~Leibs, R.~Wheeler, and
  A.~Y. Ng, ``Ros: an open-source robot operating system,'' in \emph{ICRA
  workshop on open source software}, vol.~3, no. 3.2, 2009, p.~5.

\bibitem{chitta2012moveit}
S.~Chitta, I.~Sucan, and S.~Cousins, ``Moveit![ros topics],'' \emph{IEEE
  Robotics \& Automation Magazine}, vol.~19, no.~1, pp. 18--19, 2012.

\bibitem{rusu2010fast}
R.~B. Rusu, G.~Bradski, R.~Thibaux, and J.~Hsu, ``Fast 3d recognition and pose
  using the viewpoint feature histogram,'' in \emph{IEEE/RSJ Int. Conf.
  Intelligent Robots and Systems}, 2010, pp. 2155--2162.

\bibitem{alexandre20123d}
L.~A. Alexandre, ``3d descriptors for object and category recognition: a
  comparative evaluation,'' in \emph{Workshop on Color-Depth Camera Fusion in
  Robotics at the IEEE/RSJ Int. Conf. Intelligent Robots and Systems (IROS),
  Vilamoura, Portugal}, vol.~1, no.~3, 2012, p.~7.

\bibitem{rusu2009fast}
R.~B. Rusu, N.~Blodow, and M.~Beetz, ``Fast point feature histograms (fpfh) for
  3d registration,'' in \emph{IEEE Int. Conf. Robotics and Automation}, 2009,
  pp. 3212--3217.

\bibitem{clarisse1986vicon}
O.~Clarisse and S.-K. Chang, ``Vicon,'' in \emph{Visual languages}, 1986, pp.
  151--190.

\bibitem{dna2014_tcst}
D.~Navarro-Alarcon, Y.-H. Liu, J.~G. Romero, and P.~Li, ``Energy shaping
  methods for asymptotic force regulation of compliant mechanical systems,''
  \emph{IEEE Trans. Control Syst. Technol.}, vol.~22, no.~6, pp. 2376--2383,
  2014.

\end{thebibliography}

\end{document}